\documentclass[a4paper]{llncs}

\usepackage{amssymb}
\usepackage{graphicx}
\usepackage{epstopdf}
\usepackage{color}
\usepackage{bm}
\usepackage{upgreek}
\usepackage{hyperref}
\usepackage{amsmath }

\usepackage{url}
\urldef{\mailsa}\path|{amirj,az}@robots.ox.ac.uk|
\urldef{\mailsb}\path|timor.kadir@plexalis.com|
\urldef{\mailsc}\path|emma.clark@bristol.ac.uk|

\usepackage{caption,setspace} 
\usepackage{bm}
\usepackage{amsmath}

\usepackage{algorithm}
\usepackage{program}
\usepackage{multirow}
\usepackage{booktabs}
\begin{document}

\mainmatter
\title{Predicting Spine Geometry and Scoliosis from DXA Scans}

	\author{Amir Jamaludin\inst{1}\and Timor Kadir\inst{2}\and Emma Clark\inst{3}\and Andrew Zisserman\inst{1}}
	\institute{Visual Geometry Group, Department of Engineering Science, University of Oxford
		\and
		Plexalis
		\and
		Musculoskeletal Research Unit, School of Clinical Sciences, University of Bristol\\
		\mailsa\\
		\mailsb\\
		\mailsc\\}
	\maketitle
	
	\begin{abstract} 
Our objective in this paper is to estimate spine curvature in DXA
scans. To this end we first train a neural network to predict the middle
spine curve in the scan, and then use an integral-based method to determine
the curvature along the spine curve. We use the curvature to compare
to the standard angle scoliosis measure obtained using the DXA
Scoliosis Method (DSM). The performance improves over the prior work
of Jamaludin {\it et al.} 2018.
We show that the maxium curvature can be used as a
scoring function for ordering the severity of spinal deformation.
\end{abstract}
	
	\section{Introduction}
	Scoliosis is a disease that appears as an abnormal sideways curvature of the spine often presenting in childhood and affecting up to $3\%$ of children \cite{Clark16}. In its severe form, the disease can cause lifelong disability and pain, however most cases of mild scoliosis present no symptoms and stabilize over time \cite{Asher06,Pehrsson91}. This uncertainty over whether an initial mild curve will stabilize, resolve or progress presents some challenges for effective clinical management. Therefore, our long-term goal is to develop new software tools to assist clinicians in managing such patients, and in particular to predict prognosis.
	
X-ray imaging is the standard for diagnosing and monitoring scoliotic
patients, however, since the disease presents in childhood, the
radiation burden is far from ideal, especially if used repeatedly. An
alternative approach is to use DXA imaging, which requires a far low dosage of radiation. DXA is more commonly used for measuring bone mineral density (BMD) when osteoporosis is suspected, and occasionally used to detect vertebral fractures (\cite{Bromiley15}\cite{Bromiley16}) but recently it has been shown to be
an accurate method for diagnosis of scoliosis~\cite{Taylor13};  and techniques to
automate the technique of~\cite{Taylor13} have been proposed~\cite{Jamaludin18a}. 
	
In order to assess whether a patient's disease is
progressing, stabilising or resolving it is necessary to measure the
scoliotic curvature accurately.
However,  \cite{Taylor13} focussed on the problem of the binary
classification task where patients are identified as having scoliosis
or not,  measured by as at least one curve having a {\em scoliotic angle}  of greater than 6
degrees. This angle threshold was also adopted in~\cite{Jamaludin18a}. 
The original motivation for introducing the threshold was as a conservative allowance for the fact the
DXA scans are taken with the patient lying down, rather than standing, so that the curvature of the spine
may be reduced.
However, there  is 
a growing interest in so-called
“micro-curves”, that is scoliotic curves that are below the 6
degree threshold  but might be very early
indications of problematic spines.
	
\begin{figure}[h!]
	 	\centering
	 	\includegraphics[width=1.0\textwidth]{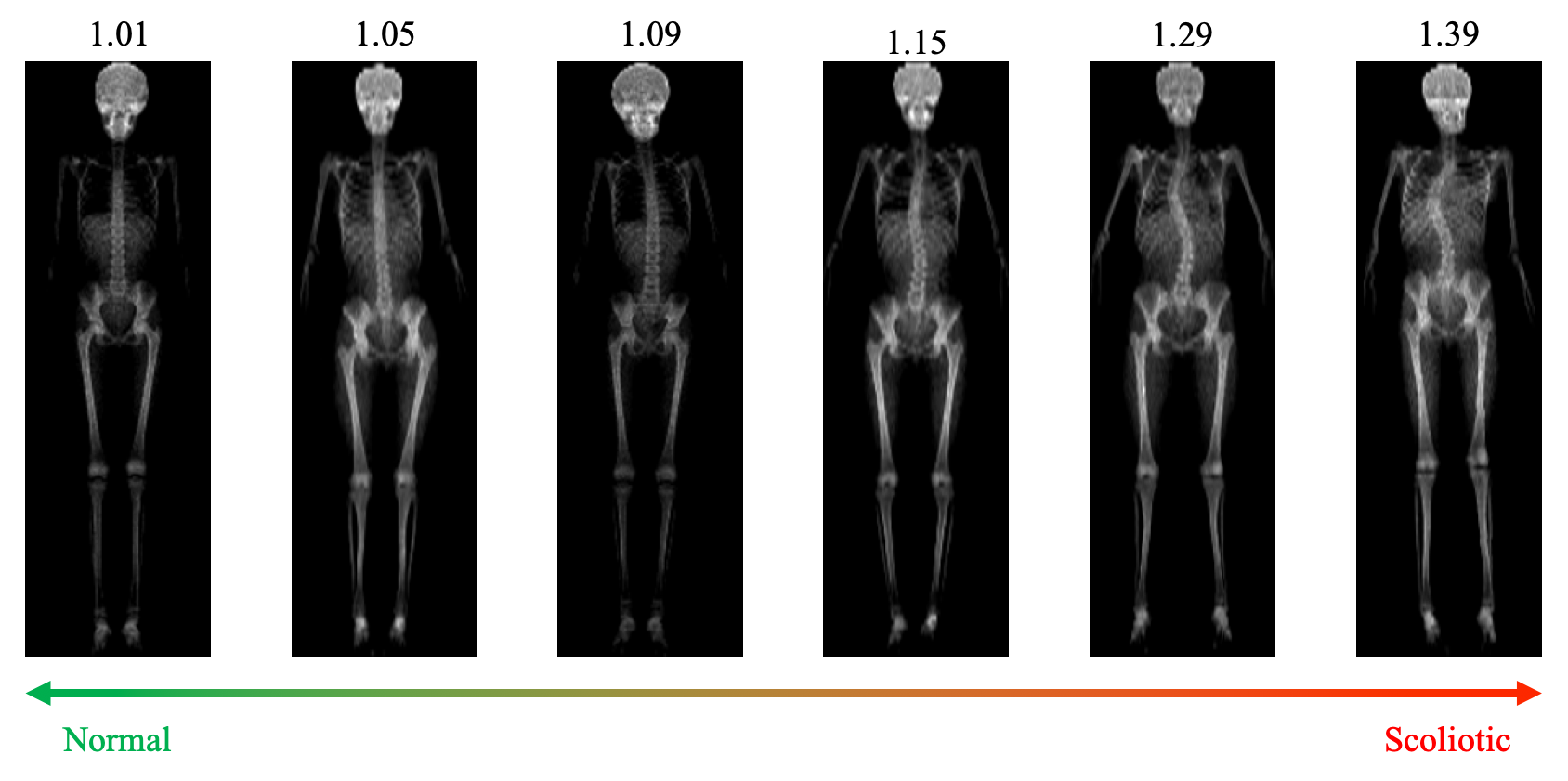}
\caption{\textbf{Severity of scoliosis:} DXA scan samples ordered by
maximum curvature of the spine i.e.\  the number on top of each scan. Scans with curvature around 1 are at the normal end of the spectrum,
while  curvatures above 1 and increasing are more scoliotic. In this example, the three samples on the left are normal scans and the three  samples on the right have scoliosis.}
	 	\label{fig:scosever}
	 \end{figure}

In this paper, we extend the work of \cite{Jamaludin18a} and make a
number of contributions. First, we propose a novel algorithm to
accurately predict the {\em curvature}  and angle  of the spine curve. 
This involves a combination of 
a state-of-the-art deep learning architecture with methods from
classical integral geometry.
We develop and
validate the algorithm on one of the largest DXA databases available
comprising 7,645 subjects, and
compare the performance
to expert defined ground-truth,  and to two alternative
baseline algorithms.
Finally, we show that the resulting curvature prediction can be used to define a score function for ordering 
severity of scoliosis in DXA scans. An example is shown in Fig.~\ref{fig:scosever}.

In the following we describe  several possible methods to predict severity of scoliosis
from DXA: (i) a baseline method to  directly regress the
angle using a CNN (Section~\ref{sec:scolioisis}); (ii) predicting the
angle by an automated version of the manual method of~\cite{Taylor13}
(Section~\ref{sec:geo});  and (iii) measuring curvature of the spine
(Section~\ref{sec:circle}). We predict angles since angle-based
measures are widely used by the medical community for scoliosis
measurement but as aforementioned we are advocating the use of
curvature since that is the underlying symptom of scoliosis. The
dataset used is similar to that used in \cite{Jamaludin18a} and as
such we also compare the network we used in this work to predict tasks
in \cite{Jamaludin18a} such as: (i) scoliosis, (ii) body positioning
error, and (iii) number of curves. These tasks are used as pre-training
for the regression network in Section~\ref{sec:scolioisis}.

	\section{The Classification \& Angle Regression Network} \label{sec:scolioisis}

The goal is to predict from a given scan the angle of the largest curve of the spine. Since the number of scans annotated with angle measurements is small, around 927, we pre-train the regression network with several classification tasks that exist in the dataset. 
	
	\subsection{Classification as Pre-training}

\begin{figure}[h!]
	\centering
	\includegraphics[width=1.0\textwidth]{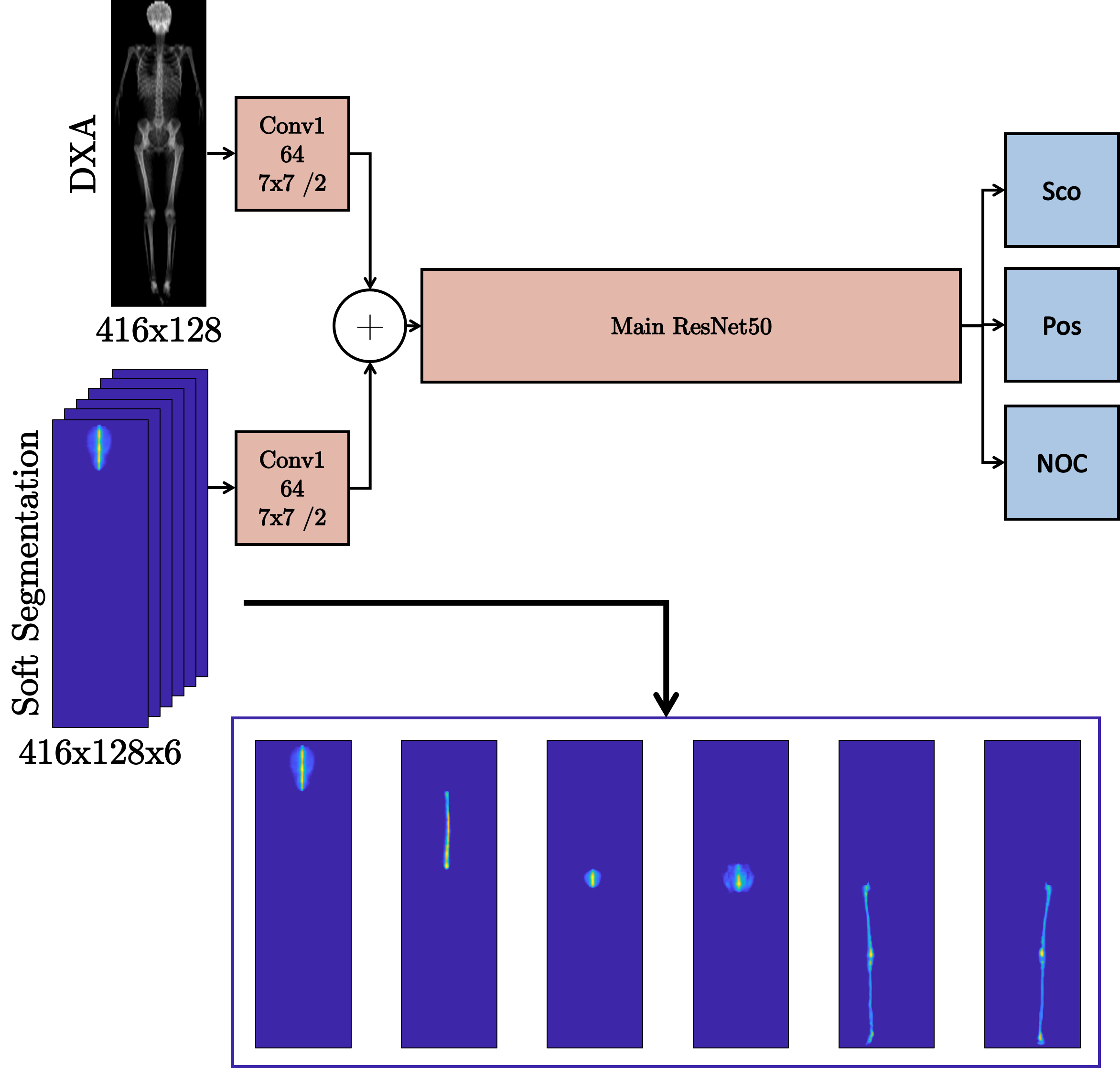}
	\caption{\textbf{Classification CNN:} The network is a modified ResNet50 \cite{He15} but with multiple inputs. The inputs are (1) raw DXA scan, and (2) soft segmentation masks where each mask corresponds to the midcurve of a specific body part. From left to right, the segmented body parts are: (1) head, (2) spine, (3) pelvic cavity, (4) pelvis, (5) right leg, and (6) left leg. The classification outputs are a binary scoliosis prediction (Sco), a binary positioning error prediction (Pos), and a multi-class prediction of the number of curves (NOC).}
	\label{fig:scocnn}
	\end{figure}

We pre-train the network on three different classification tasks: (i)
a binary classification of scoliosis vs.\ non-scoliosis, where an
angle of $6^\circ$ and above is labelled as scoliosis and vice versa,
(ii) a binary classification of body positioning error which is
dependant on the straightness of the whole body in the DXA scan, and
(iii) the number of curves of a scoliotic spine (only on cases with
scoliosis). The number of curves is divided into three different
classes: no curve (normal spine); one curve, i.e.\ a ``C'' shaped
spine; and more than one curve, which includes the classical ``S''
shaped spine with two curves or more. As noted in the introduction, the  $6^\circ$ binary cut-off was suggested in several works using DXA to measure scoliosis namely in \cite{Taylor13}, \cite{Clark14}, and \cite{Clark16}. The network is based on a
ResNet-50 but takes in two distinct inputs: (i) the raw DXA scan, and
(ii) the  soft segmentation mask produced by the segmentation network
of~\cite{Jamaludin18a}. The soft segmentation mask outputs for
each row or scanline, the score for which each pixel is most likely to
belong to the middle point of one of six body parts.
The two input streams are merged after Conv1 via addition. See Fig.~\ref{fig:scocnn}
for an overview of the network
	
	\subsubsection{Classification loss:}
	As in \cite{Jamaludin18a}, we use a multi-task balanced loss which can be expressed as minimizing a combination of the softmax log-losses of the three classification tasks:
	\begin{equation}
	\mathcal{L}_t = -\sum_{n=1}^{N}\left(y_c(x_n) - \log\sum_{j=1}^{C_t} e^{y_j(x_n)}\right)
	\label{eqn:lossTask}
	\end{equation}
	where $t$ corresponds to each classification task, with $t \in \{1 \dots 3\}$, $x$ is the input scan, $C_t$ which corresponds to the number of classes in task $t$, $y_j$ is the $j^{th}$ component of the classicication output, and $c$ is the true class of $x_n$. The loss for each classification is also balanced with the inverse of the frequency of the class to emphasize the contribution of the minority class e.g.\  only $8\%$ of the scans have scoliosis.

	\subsection{Regression} \label{subsec:reg}

The same network architecture is used for regression of the angle,
where now instead of three classification outputs in
Fig.\ref{fig:scocnn}, we have two outputs for the regression -- the angle and its uncertainty. 

Following~\cite{Novotny18c}, in training we assume a  Laplace
distribution of the measurements with a 
regression loss that is the negative log-likelihood of the Laplace distribution:
	\begin{equation}
	\mathcal{L} = \sum_{n=1}^{N}\left(-\ln \frac{1}{2 \sigma^2_n} exp\left(-\frac{\lvert y_n - \hat{y}_n \rvert}{\sigma^2_n}\right)\right)
	\label{eqn:lossReg}
	\end{equation}
where for each sample $n$ we predict both the target angle $\hat{y}_n$ and the uncertainty $\sigma^2_n$; $y_n$ in the ground truth angle. To ensure positivity of $\sigma^2_n$, we employ a softplus non-linearity to the output of the network.

	\section{Predicting the Scoliotic Angle via Geometry} \label{sec:geo}

\begin{figure}[h!]
		\centering
		\includegraphics[width=1.0\textwidth]{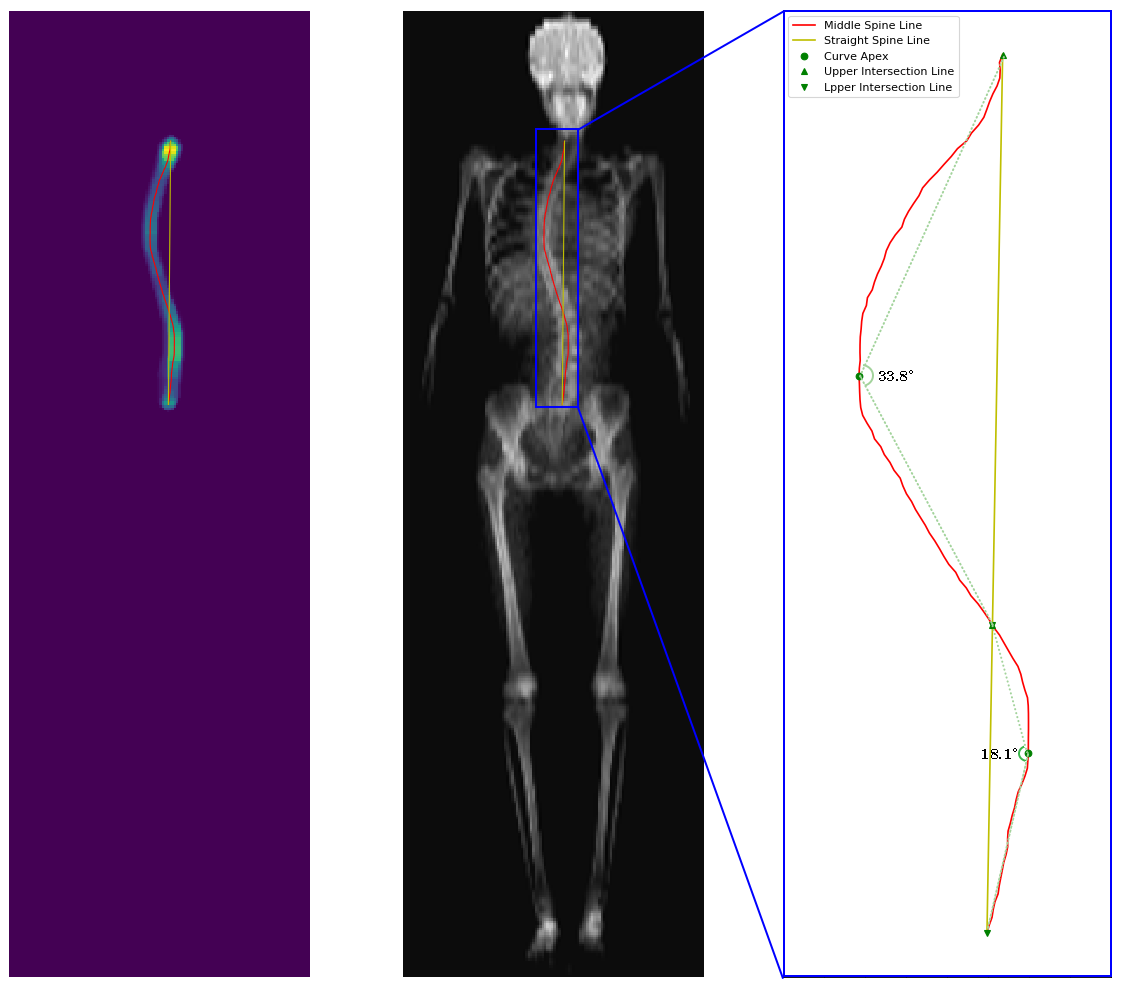}

\caption{Spine angle construction. Left:  example of the soft segmentation of the spine overlaid with the
predicted normal spine line (yellow) and middle  spine curve (red). Middle:
the spine line and curve overlaid on top of the actual DXA scan. Right: close
up view of the spine line and curve now with the intersections; there are
three intersections resulting in two curves. The ground truth angle
for this particular case is $35^\circ$, while the measured angles are
$33.8^\circ$ and $18.1^\circ$.}  \label{fig:angledsm} \end{figure}

The standard way to measure scoliosis in whole body DXA was first
introduced in \cite{Taylor13} and was called the DXA scoliosis method
(DSM). The angle measurement part of DSM is a modified Ferguson method
since the standard Cobb method cannot be used due to the low
resolution of the DXA scans. We automate this process by following and
modifying DSM, as illustrated in Fig.~\ref{fig:angledsm}.
	
\subsubsection{Normal spine line:}
\begin{figure}[h!]
  		\centering
  		\includegraphics[width=0.9\textwidth]{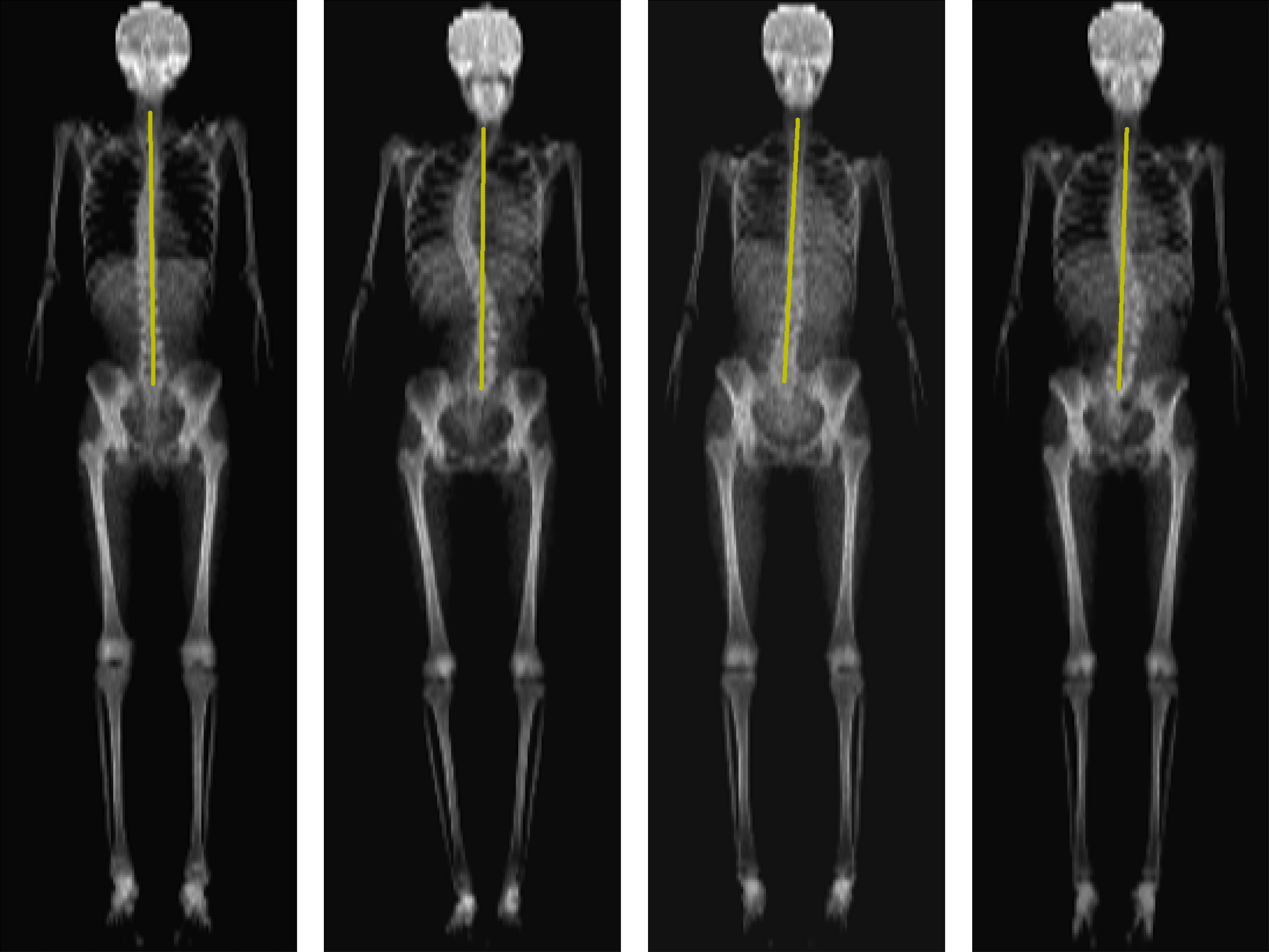}
  		\caption{Automated normal spine line obtained from the soft segmentation.}
  		\label{fig:nsl}
  	\end{figure}

The method starts by identifying the normal spine line which acts as a
reference line to measure the actual spine curve against. In
\cite{Taylor13}, the normal spine line must cross the centre point of
the spine at the level of the first rib and ends at the centre of the
spine at the fifth lumbar vertebra (L5). We approximate the spine curve by
using the soft segmentation produced in \cite{Jamaludin18a}. Each row
in the soft segmentation output of the network is a probability map of
where the midpoint of a certain body part is; the pixel with the
highest score is the predicted midpoint (see
Fig.~\ref{fig:scocnn}). The automated normal spine line is drawn from
the midpoints of the $3^{rd}$ and $97^{th}$ percentiles of the soft
segmentation mask for the spine. Fig.~\ref{fig:nsl} shows the
automated normal spine line.
  	
\subsubsection{Measuring the angle:}
The soft segmentation output is used to draw the predicted middle
spine curve. For each middle spine curve, the apex of a curve is defined
as the point furthest away from the line which lies in between two
intersections of the middle spine curve and the normal spine line. The
angle is predicted as the inner angle of the apex minus
$180^\circ$. There can exist multiple curves for a given scan, only
the maximum angle is used.
	
	\section{Measuring Curvature via an Integral method} \label{sec:circle}	

\begin{figure}[h!]
		\centering
		\includegraphics[width=1.0\textwidth]{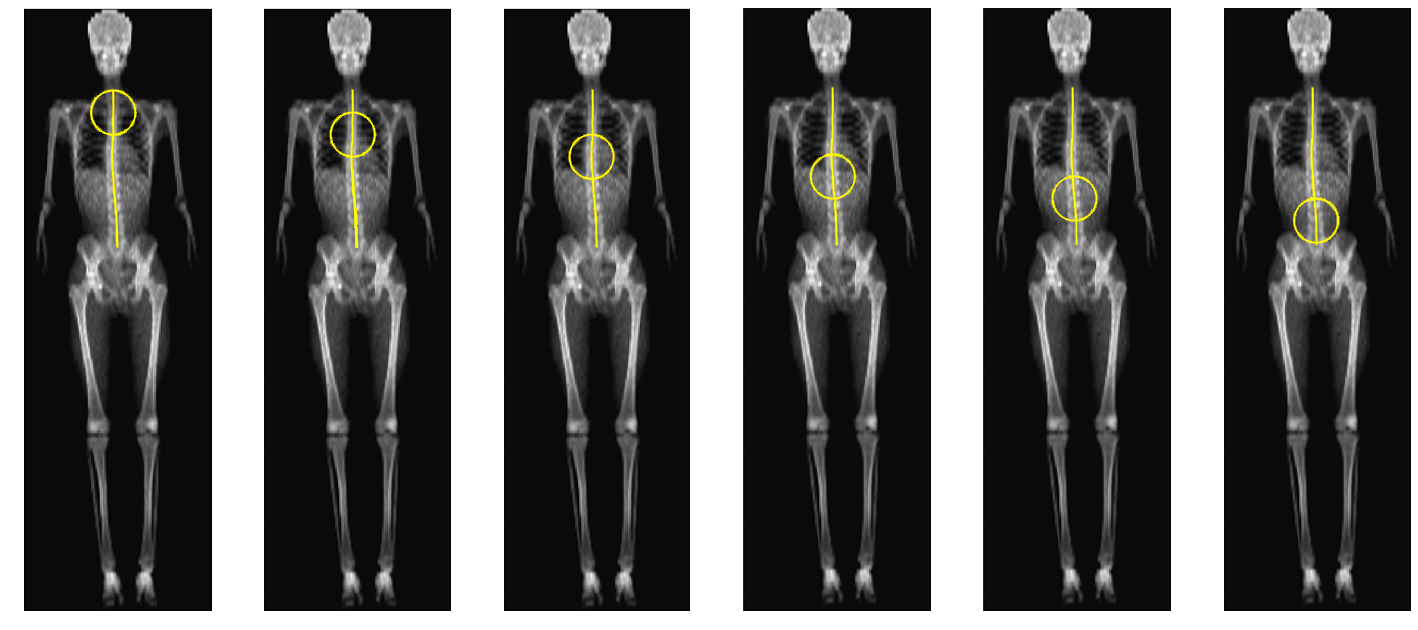}
		\caption{Curvature via area of a circle. The circle is centred on the spine curve 
and the curvature is measured via the ratio of the left and right areas of the circle. The circle is moved from the top part of the spine curve to the end, and  curvature is calculated at each point.}
		\label{fig:circ2}
	\end{figure}

\begin{figure}[h!]
		\centering
		\includegraphics[width=0.5\textwidth]{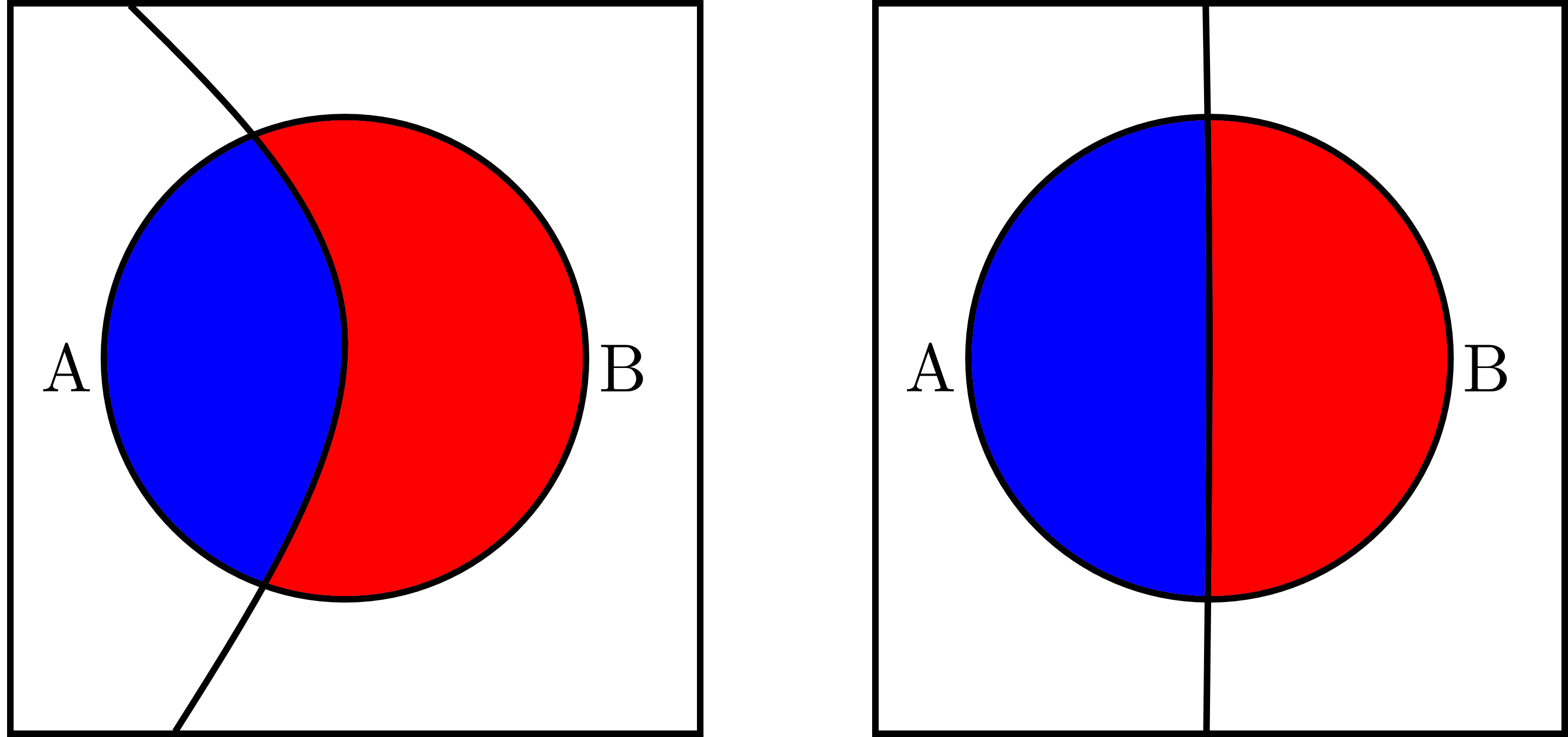}
		\caption{When the circle is centred on a curve, the areas are not equal ($Area_A \neq Area_B$), but if the circle is centred on a straight line the areas are equal  ($Area_A = Area_B$).}
		\label{fig:circ}
	\end{figure}

Scoliosis is essentially a measure of curvature of the spine and as
such we propose to directly measure curvature instead of
the angle. Measuring curvature via differential based methods is
extremely unreliable especially in our dataset. \cite{coeurjolly13}
describes an integral based curvature estimator using digital shapes;
we adapt  and simplify the method of~\cite{coeurjolly13} for curvature measurement here. In short, we
compare the two areas for a given shape e.g.\  a circle centred
on the spine curve. An example of the shape on the middle
spine curve can be seen in Fig.~\ref{fig:circ2}. The estimate of the curvature is
defined as: \begin{equation} \kappa = Area_{max}/Area_{min}
\label{eqn:curv} \end{equation}
When the spine curve is
completely straight, the ratio between the two areas is~1 
(see Fig.~\ref{fig:circ}). 
	
	\subsection{Mapping the curvature to the angle}

\begin{figure}[h!]
		\centering
		\includegraphics[width=1.0\textwidth]{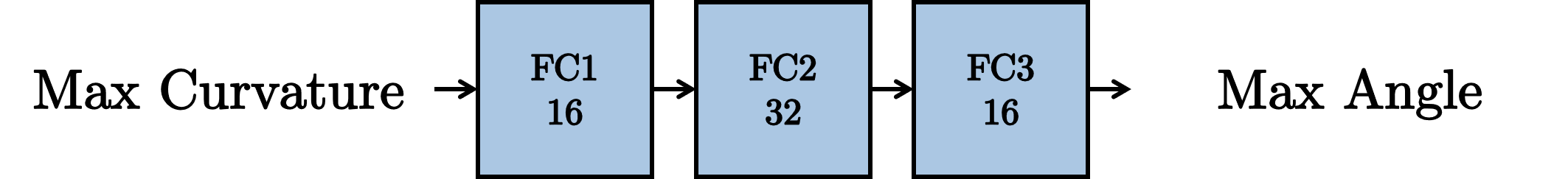}
		\caption{Network architecture that maps angle from curvature. Each fully-connected (FC) layer is followed by a ReLU activation and batch normalization.}
		\label{fig:simplereg}
	\end{figure}

Curvature is not directly comparable to the measured angle provided in the ground
truth annotation. To map the detected curvature to an angle measurement, we train
a simple fully-connected neural network regressor that takes in the
max curvature of a given spine and outputs a continuous number. We
assume that the max curvature corresponds to the max angle for a given
spine measured at the apex of the curve as in the manual method. The
regressor was trained via the same loss discussed in
Section~\ref{subsec:reg} and details of the network are given in
Fig.~\ref{fig:simplereg}.
	
	\section{Dataset \& Training Details} \label{sec:implementation}	
\begin{figure}[h!]
	\centering
	\includegraphics[width=0.9\textwidth]{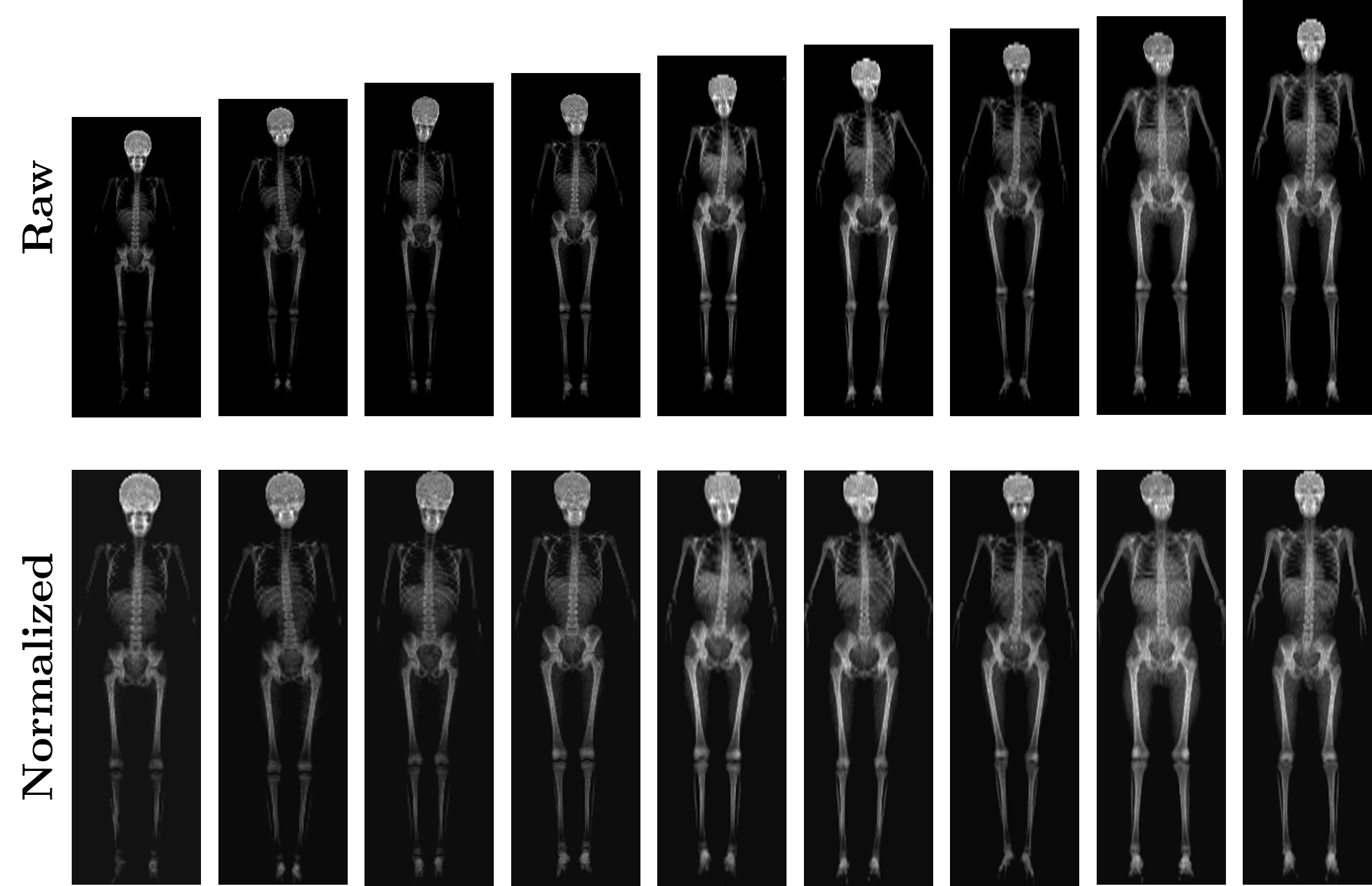}
	\caption{\textbf{Height Normalization:} The top row shows examples of scans prior to height normalization for both time points (the first five examples are from 9 year old subjects while the last four are from 15 year old subjects),  while the bottom row shows the height normalized scans.}
	\label{fig:heights}
	\end{figure}

	The dataset is from the Avon Longitudinal Study of Parents and Children (ALSPAC) cohort that recruited pregnant women in the UK. The DXA scans of the subjects were obtained from two different time points; when the subjects were 9 and 15 years of age. This difference in acquisition period and the variation of height between different individuals results  in a difference of scan heights. Fig.~\ref{fig:heights} shows the variation of scan heights in the dataset.

	In all, there are $7,645$ unique subjects in the dataset, most of which have two scans, which totals to $12,040$ scans. Most scans are annotated with several annotations which include labels such as body positioning error, and angle measurement of the curves of the spine. Angle measurements were only annotated for scans with curves deemed to be close to scoliotic or larger. The label distribution of the different classification tasks is given  in Table \ref{table:dist} while the frequency of the angle of the biggest curve of the spine is shown in Table \ref{table:freq}. We use a 80:10:10 (train:test:validation) random split to train the CNN both for the pre-training on the classification task and the angle regression, on a per patient basis (about 9.6k:1.2k:1.2k scans). A single subject and all of its scans can only appear in one of the training, validation or test  sets. Two different random splits were used and we show the mean and standard deviation of the performance at test time for the two splits.
 
	\begin{table}
		\centering
		\begin{tabular}{c|c|c|c|}
			\cline{2-4}
			& Normal                                                    & \multicolumn{2}{c|}{Abnormal}                                                                                       \\ \hline
			\multicolumn{1}{|c|}{Positioning} & \begin{tabular}[c]{@{}c@{}}10147\\ (84.3$\%$)\end{tabular} & \multicolumn{2}{c|}{\begin{tabular}[c]{@{}c@{}}1893\\ (15.7$\%$)\end{tabular}}                                      \\ \hline
			\multicolumn{1}{|c|}{Scoliosis}   & \begin{tabular}[c]{@{}c@{}}9563\\ (92.0$\%$)\end{tabular} & \multicolumn{2}{c|}{\begin{tabular}[c]{@{}c@{}}814\\ (8.0$\%$)\end{tabular}}                                        \\ \hline
			& 0                                                         & 1                                                       & \textgreater{}1                                           \\ \hline
			\multicolumn{1}{|c|}{NOC}         & \begin{tabular}[c]{@{}c@{}}9435\\ (91.1$\%$)\end{tabular} & \begin{tabular}[c]{@{}c@{}}768\\ (7.4$\%$)\end{tabular} & \begin{tabular}[c]{@{}c@{}}159 \\ (1.5 $\%$)\end{tabular} \\ \hline
		\end{tabular}
		\caption{\textbf{Distribution of classification labels:} There are three different classification tasks: (i) scoliosis, (ii) positioning error, and (iii) number of curves (NOC). There are 12,040 scans but fewer labels, since not all scans have labels for all three  tasks.}
		\label{table:dist}
	\end{table}	

\begin{table}
	\centering
	\begin{tabular}{|l|c|c|c|c|c|c|c|c|c|c|c|c|c|c|}
		\hline
		Angle     & 0$^\circ$ & 1$^\circ$ -- 4$^\circ$ & 5$^\circ$ & 6$^\circ$ & 7$^\circ$ & 8$^\circ$ & 9$^\circ$ & 10$^\circ$ & 11$^\circ$ & 12$^\circ$ & 13$^\circ$ & 14$^\circ$ & 15$^\circ$ & 16$^\circ$+ \\ \hline
		Frequency & 9450      & 28                     & 85        & 116       & 146       & 144       & 113       & 81         & 60         & 45         & 18         & 16         & 12         & 63          \\ \hline
	\end{tabular}
	\caption{\textbf{Distribution of annotated angle in the dataset:} There are 12,040 scans but only 927 were annotated; any scan annotated as $6^\circ$ and above are labelled as scoliosis. Cases marked as $0^\circ$ are unreliable as they may actually have a small curvature, but since they were deemed to be below the scoliosis cut-off, they were marked down as $0^\circ$ during the annotation process. As such, only scans with $>0^\circ$ are used in our validation experiments. Only the angle of the largest curve for a given scan is shown here; a scan can have multiple curves. In all, we have 814 scoliotic and 113 non-scoliotic cases.}
	\label{table:freq}
\end{table}	

	\subsubsection{Pre-processing:} 
	The scans are normalized such that both the head and feet are roughly in the same region for all the scans regardless of age and original height of the scans. First, empty spaces on top of the head and below the feet are also removed. Then, the scans are resized and cropped isotropically to prevent distortion and to keep the aspect ratio the same as the original. The dimensions of the scans after normalization is $416 \times 128$ pixels while the raw dimensions of the scans vary from $173 \times 128$ to $411 \times 128$ pixels.
	
	\subsubsection{Training Details:} Both the classification and regression networks are optimized via Adaptive Moment Estimation (Adam) from scratch. The hyperparameters are; batch size of 128; beta1 0.9; beta2 0.999; initial learning rate is 0.0001 and lowered by a factor of 10 as the loss plateaus. The network were trained via PyTorch using an NVIDIA Titan X GPU. We employ several training augmentation strategies: (i) translation of $\pm24$ pixels in the x-axis, (ii) translation of $\pm24$ pixels in the y-axis, and (iii) random horizontal flipping. At test time, the final prediction is calculated from the average prediction of an image and its flip.

	\section{Experiments \& Results}\label{sec:exp}

We compare all the methods to predict angles from DXA scans which includes the regression network in Section \ref{sec:scolioisis}, the automated automated DSM in Section \ref{sec:geo}, and the curvature via integral geometry in Section~\ref{sec:circle}. 

	\subsection{Classification} 

First, since we pre-trained the regression network in Section \ref{sec:scolioisis} to classify certain tasks, we compare the performance to existing literature. We report better performance owing to the deeper and modern network
used,  i.e.\ the ResNet50 \cite{He15}, against the VGG-M \cite{Chatfield14} style network used in
\cite{Jamaludin18a}. See Table \ref{table:class}. The biggest
improvement is in the prediction of the number of curves $72.6\%
\rightarrow 77.3\%$ which might be correlated with the improvement of
scoliosis prediction $90.5\% \rightarrow 94.2\%$ as these two tasks
are closely linked i.e.\ more pronounced scoliosis or curvature of the
spine normally appear with more than one curves.
		
	\begin{table}[h!]
	\centering{
		\begin{tabular}{l|l|l|}
			\cline{2-3}
			& \cite{Jamaludin18a} & ResNet50       \\ \hline
			\multicolumn{1}{|l|}{Scoliosis}        & $90.5 \pm 1.5$                       & $94.2 \pm 2.1$ \\ \hline
			\multicolumn{1}{|l|}{Body Positioning} & $80.5 \pm 0.3$                       & $81.5 \pm 1.1$ \\ \hline
			\multicolumn{1}{|l|}{Number of Curves} & $72.6 \pm 1.2$                       & $77.3 \pm 1.4$ \\ \hline
		\end{tabular}
	}
	\caption{\textbf{Average Per-class Accuracy (mean $\mathbf{\pm}$ std $\mathbf{\%}$):} We compare against \cite{Jamaludin18a}.}
	\label{table:class}
	\end{table}	
	
	\subsection{Comparing Angle Measurements} 

\begin{figure}[h!]
		\centering
		\includegraphics[width=1.0\textwidth]{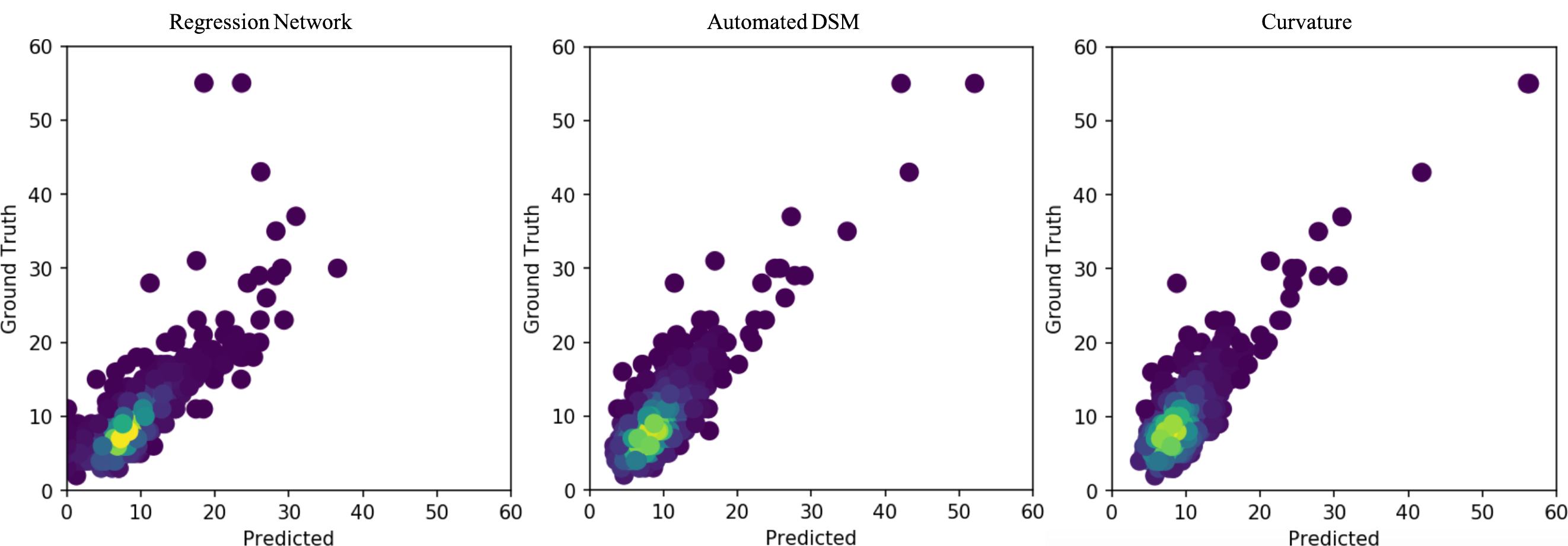}
		\caption{Scatter plot of all the angle annotated versus the three methods.}
		\label{fig:scatter}
	\end{figure}

Fig.~\ref{fig:scatter} shows scatter plots of the ground truth
against the proposed methods.
Surprisingly, the correlation between the ground truth annotation and predictions is similar
for three different methods. The correlation is: 0.79 for the regression
network; 0.82 for the automated DSM; and 0.82 for the curvature method.
However, it can be seen that the regression network (Section \ref{sec:scolioisis})
struggles to predict extreme scoliosis, $>40^\circ$, and constantly over-predicts 
$0^\circ$; possibly due to the abundance of cases with micro-curves that were graded as ``normal'' during pre-training.  The other methods, namely automated DSM and curvature, do not suffer from this problem.

\begin{figure}[h!]
		\centering
		\includegraphics[width=0.8\textwidth]{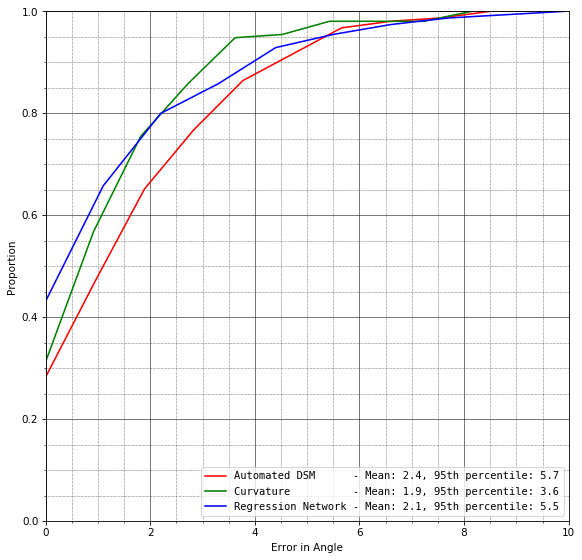}
		\caption{Comparing the three methods against ground truth on the test set. The x-axis is the error in terms of angle between the prediction and the ground truth,  while the y-axis is the proportion of the test set. The curvature method works best with 95$\%$ of the data having at most $3.6^\circ$ error.}
		\label{fig:compareanglepred}
	\end{figure}

	We also compare the angle prediction of each method against the ground truth angle in Fig.~\ref{fig:compareanglepred}.

The average error of all the methods are quite similar
with: the curvature-based method  (mapping to the angle via a neural network)
 having an average error of  just $1.9^\circ$; the regression
network having  $2.1^\circ$;  and the automated DSM having
$2.4^\circ$. However, looking at proportion of error we can see a
clear difference where $95\%$ of the data fall below $3.6^\circ$ error
for the curvature-based method while the error is $\gt5.5^\circ$ for
the other methods.

 For comparison, a manual method using ultrasound imaging to measure coronal curvature in subjects with scoliosis reported intra-rater correlation coefficients ranging from $0.84$ to $0.93$ with the standard error ranging from $1.6^\circ$ to $2.8^\circ$ \cite{Zheng15}.
	
	\subsection{Qualitative Evaluation} 
	\subsubsection{Curvature Heatmaps.}
\begin{figure}[h!]
	\centering
	\includegraphics[width=0.8\textwidth]{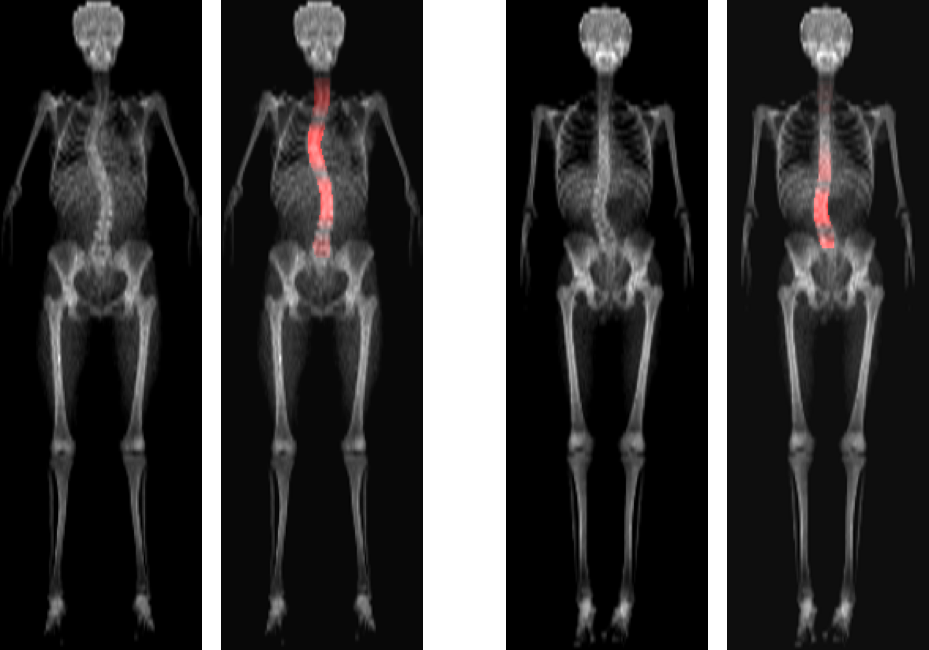}
	\caption{\textbf{Scoliosis Heatmaps:} The left pair shows thoracic scoliosis,  while the right shows lumbar scoliosis.
Each pair includes the input image  and the image with the heatmap overlaid.}
	\label{fig:hotspots}
	\end{figure}

	Unlike a CNN as in \cite{Jamaludin18a} that can produce evidence hotspots, we instead produce heatmaps of curvature by mapping the values to the segmentation mask on a per scanline basis. This is done since there is no CNN involved in producing the curvature of the line. The segmentation is the same soft segmentation used in Section \ref{sec:scolioisis}; examples can be seen in Fig.~\ref{fig:scocnn}. We show examples of scoliotic scans alongside their heatmaps in Fig.\ref{fig:hotspots}. As expected, the heatmaps are brighter on regions with high curvature and vice versa. In Fig.\ref{fig:hotspots}, we can see that these heatmaps highlight specific regions according to the type of scoliosis i.e.\  a spine with thoracic scoliosis is brighter around the thoracic region (upper spine) and similarly a spine with lumbar scoliosis is brighter around the lumbar region (lower spine).
	
	\subsubsection{Severity of Scoliosis.}
	The maximum curvature can be interpreted as a form of scoliosis score; since the curvature directly relates to the shape of the spine, the larger the curvature the more likely that the scan has scoliosis. Fig.\ref{fig:scosever} shows scans on the test set alongside their curvature. This curvature can be used to monitor disease progression of patients with scoliosis, where an increase of the curvature across a period of time,  i.e.\  a longitudinal study of the subject,  would mean the scoliosis is getting worse.

	\section{Conclusion}
	We have shown that there is a correlation between the curvature of the spine measured automatically in DXA scans and the angles measured by clinical experts. We have also shown that measuring curvature is slightly more beneficial in terms of regression performance in terms of angle and that we can reliably use these detected curvature values to represent how scoliotic a spine is in DXA.
	
	{\small \subsubsection*{Acknowledgements.}  We are extremely grateful to all the families who took part in this study, the midwives for their help in recruiting them, and the whole ALSPAC team, which 	includes interviewers, computer and laboratory technicians, clerical workers, research scientists, volunteers ,managers, receptionists and nurses. The	UK Medical Research Council and the Wellcome Trust (Grant ref: 102215/2/13/2) and the University of Bristol provide core support for ALSPAC. This publication is the	work of the authors and Amir Jamaludin will serve as guarantor for the contents of	this paper. This research was specifically funded by the British Scoliosis Research 	Foundation, and the DXA scans were funded through the Wellcome Trust (grants 084632	and 079960).}
	
	\bibliographystyle{splncs03}
	\bibliography{Bib/shortstrings,Bib/vgg_local,Bib/vgg_other,Bib/mybib}
	\end{document}